\documentclass[11pt]{article}
\usepackage{subcaption}
\usepackage{amsmath,amsfonts,amssymb}
\usepackage{graphicx}
\usepackage{setspace}
\usepackage{csquotes}
\usepackage{hyperref}

\title{On the Logic Elements Associated with Round-Off Errors and
  Gaussian Blur in Image Registration:  A Simple Case of Commingling}

\author{Serap A.~Savari \\ Texas A\&M University, College Station, TX 77843-3128, USA}

\begin{document}

\maketitle

\begin{abstract}
  Discrete image registration can be a strategy to reconstruct signals
  from samples corrupted by blur and noise.
 We examine superresolution and discrete image registration for
  one-dimensional spatially-limited piecewise constant functions which are
  subject to blur which is Gaussian or a mixture of Gaussians as well as to
  round-off errors.  Previous approaches address the signal recovery problem
  as an optimization problem.  We focus on a regime with low blur and
  suggest that the operations of blur, sampling, and quantization are not
  unlike the operation of a computer program and have an abstraction
  that can be studied with a type of logic.  When the minimum distance
  between discontinuity points is between $1.5$ and 2 times the sampling
  interval, we can encounter the simplest form of a type of interference
  between discontinuity points that we call ``commingling.''  We describe
  a way to reason about two sets of samples of the same signal that
  will often result in the correct recovery of signal amplitudes.
  We also discuss ways to estimate bounds on the distances between
  discontinuity points.
\end{abstract}

\section{Introduction and Motivating Example}
One of the foundational topics in the study of image data is the recovery
of signals from samples distorted by blur and noise.
This problem has resulted in numerous contributions pertaining to the
computational superresolution of data from 
instruments or sensors with limited resolution 
(see, e.g., \cite{tong}. \cite{candes},
\cite{mri}, \cite{fw} and the references therein).
It is also related to a large literature in image registration, which
combines information from two or more images of the same signal
\cite{cheng}, \cite{tong}, \cite[\S 2.5]{ip}.
Digital images are quantized \cite[\S 2.4]{ip}, but previous studies
do not treat that feature as fundamental to the understanding of blur
and instead typically focus on optimization (\cite{candes},
\cite{mri}, \cite{fw}, \cite[\S 5.9]{ip}).
Digital images are also sampled \cite[\S 2.4]{ip}.
Given the importance of the applications that involve image data,
it is worth asking if disciplines that primarily consider the processing
of discrete data have the potential to advance image analysis, and we will
use a simple special case of the problem to suggest that they do.
For example, signal recovery algorithms must be implemented in hardware
and/or software to be applied.
There is a significant body of work on the logic underlying
computer programs and circuits (see, e.g., \cite{asu}, \cite{hoare},
\cite{milner}, \cite{d}).
In this work we begin to incorporate the philosophy of sequential and
concurrent processing to the understanding of the signal recovery
problem; we focus in this paper exclusively on the effects of blur,
sampling, and round-off errors on one-dimensional piecewise constant
functions under simple conditions, and we only briefly comment on
stochastic noise.

We motivate our investigations with an example.
Suppose our uniform sampling interval is $T$ and our signal is
  \begin{displaymath}
g(t) \; = \; \left\{ \begin{array}{ll}
1, & 0 \leq t < 1.51T, \; 3.02T \leq t < 4.53T \\
-1, & 1.51T \leq t < 3.02T, \; 4.53T \leq t < 6.04T \\
0, & \mbox{otherwise.}
\end{array}
\right.
  \end{displaymath}

  Our focus throughout will be on blur which is Gaussian or a mixture of
  Gaussians \cite[\S5.6]{ip}.  For our example we will consider pure
  Gaussian blur of the form
  $$   h(t)  = \frac{e^{-\frac{t^2}{2\sigma^2}}}{\sigma \sqrt{2 \pi}} , \; - \infty
  < t < \infty . $$
Then $\tilde{g} (t) = g(t) \ast h(t)$ is a blurred version of $g(t)$ without
round-off errors.  We observe eleven samples of this blurred version
which are each rounded to the nearest integer multiples of
$\frac{1}{256}$.  For our first set of samples, suppose $\sigma =
\frac{T}{8}$ and we begin sampling at -1.8T.  Then our first set of
observations is
$$    \gamma_0  =  \left( 0 , \ 0 , \ \frac{242}{256} , \ \frac{253}{256} , \
    -1, \ \frac{218}{256} , \ \frac{254}{256} , \ -1, \ -\frac{26}{256} , \
    0, \ 0 \right) $$ .

    Notice that we cannot recover $g(t)$ from $\gamma_0$.
    For example, we again obtain $\gamma_0$ after rounding if we
    apply Gaussian blur with $\sigma = \frac{T}{7.7}$ to the function
  \begin{displaymath}
g(t) \; = \; \left\{ \begin{array}{ll}
\frac{129}{128}, & 0 \leq t < 1.508T, \\
-1, & 1.508T \leq t < 3.014T, \; 4.527T \leq t < 6.035T \\
\frac{257}{256}, & 3.014T \leq t < 4.527T \\
0, & \mbox{otherwise.}
\end{array}
\right.
  \end{displaymath}

  The application of imag registration \cite{tong} suggests that a second
  set of samples may be helpful.  Suppose the blur is now
  $\sigma = \frac{T}{7}$ and we begin sampling at -1.3T.
Our second set of observations is
$$ \gamma_1  =  \left( 0 , \ \frac{5}{256} , \ 1, \ -\frac{209}{256} , \
 -\frac{250}{256} , \ 1, \ -\frac{196}{256} , \ -\frac{254}{256} , \
 0, \   0, \ 0 \right). $$

 The question is how to combine $\gamma_0$ and $\gamma_1$.
 If we either directly average them \cite[p. 70]{ip}
 or average them after applying the maximum of cross-correlation template
 matching \cite[p. 1061]{ip}, then it is not apparent how to infer much
 about $g(t)$.  However, we will show that there is a way to reason about
 $\gamma_0$ and $\gamma_1$ to recover the amplitudes of $g(t)$.

 A spatially limited piecewise constant function with a minimum distance
 between discontinuity points can be viewed in terms of a sequence of
 events consisting of actions to modify the amplitude of the signal
 at certain (continuous) times.  These events are input to the combined,
 blur, sampling, and quantization operation which outputs partial
 information about the events of interest.  Therefore, the difference
 between successive samples provides a useful representation of the output.
 For this difference representation and the blur regime we study, there
 are three behavioral primitives.  In our earlier work
 \cite{savari}-\cite{s4}, any difference value could be affected by at
 most one discontinuity point.  However, a difference value could be
 affected by more than one discontinuity point, and we call that
 phenomenon {\em commingling}.  In this paper we consider the simplest
 form of commingling where there is exactly one way for a difference
 value to be impacted by two discontinuity points; the new behavior
 is a type of fusion between two primitive behaviors.
 
 The plan for the rest of the paper is as follows.  In Section~2, we
 review a measurement and a difference matrix from \cite{s4}.
 In Section~3 we describe the three primitive behaviors and add a fourth.
 We will also revisit our motivating example there.
 In Section~4, we give an overview of the syntax of our model.
 In Section~5, we describe an approach to parse our data and make
 inferences about signal amplitudes.
 In Section~6 we offer simple bounds on the relative locations of
 discontinuity points.
  In Section~7, we conclude.

\section{Preliminaries}
As in \cite{savari}-\cite{s3}, our underlying spatially
limited piecewise constant function 
with $m$ regions in its support by 
\begin{displaymath}
g(t) \; = \; \left\{ \begin{array}{ll}
g_j, & D_{j-1}  \leq t < D_j, \; j \in \{1, \ \dots , \ m\} \\
0, & \mbox{otherwise},
\end{array}
\right.
\end{displaymath}
where $D_0 = 0, \ g_1 \neq 0, \; g_m \neq 0,$ 
$g_j \neq g_{j+1}, \ j \in \{1, \ \dots ,
\ m-1\}$, and $g_0 = g_{m+1} = 0 $.  As in \cite{s4}, suppose that
$g_j$ is an integer multiple of $\frac{1}{256}$ with bounded magnitude
for all $j$.

We consider uniform sampling with a sampling interval of length $T$
and assume it is impossible to sample at a
discontinuity point.  Further suppose that no two discontinuity points have
a distance which is an integer multiple of $T$.

Let $h(t)$ be a pure Gaussian blur 
\begin{eqnarray*}
  h(t) & = & \frac{e^{-\frac{t^2}{2\sigma^2}}}{\sigma \sqrt{2 \pi}} , \; - \infty
  < t < \infty \\
  \mbox{and } \; \Phi (z) & = & \int_{- \infty}^{z} \frac{e^{-t^2 / 2}}{\sqrt{2 \pi}} dt.
  \end{eqnarray*}
Let $\tilde{g} (t) = g(t) \ast h(t)$ be a blurred version of $g(t)$ without
round-off errors.  As we discussed in \cite{s4}
\begin{displaymath}
  \tilde{g} (t) 
   =  \sum_{j=0}^{m} (g_{j+1}-g_j)
   \Phi \left( \frac{t-D_j}{\sigma} \right).
\end{displaymath}
As in \cite{s4},
we observe $N$ samples of $\tilde{g} (t)$ beginning at $t_0 < 0$
and ending at $t_0 +(N-1)T > D_m.$
To first characterize the unrounded samples, we follow \cite{s4} and
let $\tilde{M}$ be the $N \times (m+1)$ deformation matrix given by
\begin{displaymath}
  \tilde{M}_{i,j} \; = \; \Phi \left( \frac{t_0 + iT -D_j}{\sigma} \right),
  \; i \in \{0, \ \dots , \ N-1 \},   \; j \in \{0, \ \dots , \ m \},
\end{displaymath}
$g_D$ be the {\em difference vector} 
\begin{displaymath}
  g_D \; = \; (g_1-g_0, \ g_2-g_1, \ \dots , \ g_m-g_{m-1}, \ g_{m+1}- g_m)^T,
\end{displaymath}
$\tilde{g} [i]$ be the unrounded sample $\tilde{g} [i] = \tilde{g} (t_0 + iT), \;
  i \in \{0, \ \dots , \ N-1 \}$, and
\begin{displaymath}
  \tilde{g} \; = \; (  \tilde{g} [0], \ \tilde{g} [1], \ \dots , \
  \tilde{g} [N-1])^T .
\end{displaymath}
Then
  \begin{displaymath}
  \tilde{g} \; = \;   \tilde{M} {g}_D .
  \end{displaymath}
  In the
  absence of statistical noise, our observation of quantized samples is
  \begin{displaymath}
  \gamma \; = \; (  \gamma [0], \ \gamma [1], \ \dots , \
  \gamma [N-1])^T,
  \end{displaymath}
  where $\gamma [i]$ is the closest integer multiple of
$  \frac{1}{256}$ to $\tilde{g} [i]$.  Then
    \begin{displaymath}
  \gamma \; = \;   {M} {g}_D ,
  \end{displaymath}
    where the signal-dependent {\em measurement matrix} $M$ is a corrupted
    version of $\tilde{M}$.  Moreover, it is straightforward to generalize
    the discussion to a mixture of Gaussian blurs.
    In \cite{s4} we briefly discuss the cases of
    pure Gaussian blur with $\sigma$ extremely large or extremely small;
    in the former case each element of $M$ is $0.5$ and in the latter case each
    element of $M$ is either 0 or 1.  The main focus of \cite{s4} and this
    paper is in a regime of small but apparent blur where each column of the
    measurement matrix may have at most one element strictly between zero
    and one; these
 elements of the measurement matrix {\em critical values}.
The measurement matrix depends on the value $\nu_j$, 
$j \in \{0, \ 1, \ \dots , \ m \}$, for which
\begin{equation}
  \Phi \left( \nu_j \right)  \; = \;
    1- \frac{1}{512  |g_{j+1} - g_{j} |}. \label{eq:p0}
\end{equation}
For pure Gauusian blur, if
\begin{displaymath}
  \frac{t_0 + iT -D_j}{\sigma} \ > \ \nu_j , \; \mbox{then} \; M_{i,j} = 1,
\end{displaymath}
and if
\begin{displaymath}
  \frac{t_0 + iT -D_j}{\sigma} \ < \ - \nu_j , \; \mbox{then} \; M_{i,j} = 0.
\end{displaymath}

For a weighted average of pure Gaussian blurs of the form we study,
let $\sigma_{\max}$ be the largest $\sigma$ that contributes to the
round-off errors associated with the blur.
Then \cite[Proposition 1]{s4} is 

\noindent {\bf Proposition 1:}
For pure Gaussian blur or a mixture of Gaussian blurs,  if
\begin{equation}
  \sigma_{\max} < \frac{0.5 T}{ \max_j \nu_j}
  \end{equation}
    then each column of the measurement matrix has at most one critical value.

    The measurement matrix also depends on the number of samples taken in the
    various regions of $\tilde{g}(t)$.
    Assume $\eta_0$ samples of $\tilde{g} (t)$ are taken for $t<0$,
$\eta_j$ samples are taken in the region $D_{j-1} < t < D_j, \;
j \in \{1, \ \dots , \ m \}$ and $\eta_{m+1}$ samples are taken for
$t> D_m$.  $\eta_0$ and $\eta_{m+1}$ are each at least one,
and the constraints on $\eta_1 , \ \dots , \ \eta_m$ are discussed in
\cite{savari}.
For $j \in \{0, \ 1, \ \dots , \ m \}$, let
\begin{displaymath}
  \iota (j) \; = \; \sum_{k=0}^j \eta_k.
\end{displaymath}
Then $\gamma [\iota (j)]$ is the first sample following $D_j$, and index
$ \iota (j)$ is called a {\em segmentation point}.

In \cite{s3} and \cite{s4} we work with a {\em difference matrix} $M_D$
to address the possible variations in
$\eta_j, \; j \in \{0, \ \dots , \ m+1 \}$.  $M_D$ has the same first row
as $M$ and defines row $i$ for $i \in \{1, \ \dots , \ N-1 \}$
as row $i$ of $M$ minus row $i-1$ of $M$.
    Then \cite[Corollary 2]{s4} is

    \noindent {\bf Corollary 2:} 
Given a pure Gaussian blur or 
a mixture of Gaussian blurs with $\sigma_{\max} < 0.5 T / \max_j \nu_j$ for all
$k$, there are three possible forms for column $j$ of the measurement matrix:
\begin{itemize}
\item $M_{i,j} = 0$ for $i \leq \iota (j)-1$, and
  $M_{i,j} = 1$ for $i \geq \iota (j)$
\item $M_{i,j} = 0$ for $i \leq \iota (j)-1$, $0.5 < M_{\iota (j),j} < 1$,
  and   $M_{i,j} = 1$ for $i \geq \iota (j)+1$
  \item $M_{i,j} = 0$ for $i \leq \iota (j)-2$, $0 < M_{\iota (j)-1,j} < 0.5$,
  and   $M_{i,j} = 1$ for $i \geq \iota (j)$.
\end{itemize}
Therefore, the three possible forms for column $j$ of the difference matrix are:
  \begin{itemize}
  \item $M_{D[\iota (j),j]} = 1, \; M_{D[i,j]} = 0$ for $i \neq \iota (j)$
  \item $M_{D[\iota (j),j]} = M_{\iota (j), j} \in (0.5, \ 1), \;
    M_{D[\iota (j)+1,j]} = 1-M_{\iota (j), j}$, \\ and
    $M_{D[i,j]} = 0$ for $i \notin \{\iota (j), \ \iota (j)+1 \}.$
      \item $M_{D[\iota (j)-1,j]} = M_{\iota (j)-1, j} \in (0, \ 0.5), \;
    M_{D[\iota (j),j]} = 1-M_{\iota (j)-1, j}$, \\ and
    $M_{D[i,j]} = 0$ for $i \notin \{\iota (j) -1, \ \iota (j) \}.$
  \end{itemize}
  The samples and differences between consecutive samples depend on the
  rows of the measurement matrix.  The following result is
  \cite[Proposition 3]{s4}:

  \noindent {\bf Proposition 3:}
  For a pure Gaussian blur or a mixture of Gaussian blurs with
  $\sigma_{\max} < 0.5 T / \max_j \nu_j$,  no row of the measurement matrix
  contains more than one component
which is a critical value.  Moreover, if the minimum distance between
discontinuity points exceeds $2T$, then any row of $M_D$ has at most one
nonzero component.

The algorithms we propose for image registration in \cite{s3} and
\cite{s4} are based on $M_D g_D$. In \cite{s3} we see that when the blur
is negligible compared to quantization errors, then any
component of $g_D$ appears in
one element of $M_D g_D$ assuming a minimum distance of at least $T$ between
discontinuity points.  Proposition~3 indicates that in this regime of
larger blur, any component of $g_D$ can either appear
in one element of $M_D g_D$ or be divided between two consecutive components
of $M_D g_D$ without the commingling of different elements of $g_D$ as long
as the minimum distance between discontinuity points of $g(t)$ is at
least $2T$.  However, when the minimum distance between discontinuity points
falls below $2T$, then the commingling of different element of $g_D$
within $M_D g_D$ is possible.  Therefore, it is next of interest to consider
a situation where there is a single form of commingling that may occur.

\subsection{On Segmentation Points and Minimum Distance 1.5T}
The reason why a minimum distance of $2T$ has a special property in this
blur model can be viewed as a consequence of Corollary~2 and the following
more general result.  We defer all proofs to a longer version of the paper.

\noindent {\bf Proposition 4:}  For any positive integer $l$, if the minimum
distance between discontinuity points exceeds $\frac{(l+1) T}{l}$, then there
  is no collection of $l+1$ consecutive segmentation points.

  Just as the case $l=1$ has a special property, the case $l=2$ is the only
  one where there is exactly one type of commingling of elements of $g_D$
  that can occur in $M_D g_D$ under this blur model.  Our first step in
  the study of commingling is to extend Proposition~3.
  We have

    \noindent {\bf Proposition 5:}
  For a pure Gaussian blur or a mixture of Gaussian blurs with
  $\sigma_{\max} < 0.5 T / \max_j \nu_j$,  no row of the difference matrix
  more than two nonzero components.  Moreover, if a row of the difference
  matrix contains two nonzero values, then they will be consecutive
  components and at least one of them will be strictly between $0$ and $1$.
Furthermore, if the minimum distance between
discontinuity points exceeds $1.5T$, then for a row of $M_D$ with two
nonzero entries, both of them will have values strictly between 0 and 0.5.

We also have

    \noindent {\bf Proposition 6:}
  For a pure Gaussian blur or a mixture of Gaussian blurs with
  $\sigma_{\max} < 0.5 T / \max_j \nu_j$ and assuming the
  minimum distance between
  discontinuity points exceeds $1.5T$,  if rows $i_1$ and $i_2 > i_1$
  of the difference matrix each contain two nonzero components, then
  $i_2 \geq i_1 +3.$

  We will also use the terminology and notation of
  {\em difference sequence} for $\delta =   M_D g_D$.

  \section{Modeling and Our Motivating Example}
  By Corollary~2, since there are three possible forms for column $j$
  of the difference matrix it follows that
  there are three primitive behaviors available
  in an observation.  In \cite[Section 5]{s4} we summarize these with
  tokens that we will expand upon here.  First we ignore commingling.
     \begin{itemize}
     \item $M_{D[\iota (j),j]} = 1, \; M_{D[i,j]} = 0$ for $i \neq \iota (j)$.
       Here we denote $\iota(j)$ with the token $A$ and
       $\delta (\iota(j)) = g_{j+1} - g_j.$
  \item $M_{D[\iota (j),j]} = M_{\iota (j), j} \in (0.5, \ 1), \;
    M_{D[\iota (j)+1,j]} = 1-M_{\iota (j), j}$, \\ and
    $M_{D[i,j]} = 0$ for $i \notin \{\iota (j), \ \iota (j)+1 \}.$
    Here we label the pair $(\iota(j), \iota(j)+1)$ with the tokens
    $(F_1, \ F_2)$ and $\delta (\iota(j)) + \delta (\iota(j)+1)
    = g_{j+1} - g_j.$
      \item $M_{D[\iota (j)-1,j]} = M_{\iota (j)-1, j} \in (0, \ 0.5), \;
    M_{D[\iota (j),j]} = 1-M_{\iota (j)-1, j}$, \\ and
    $M_{D[i,j]} = 0$ for $i \notin \{\iota (j) -1, \ \iota (j) \}.$
    Here we label the pair $(\iota(j)-1, \iota(j))$ with the tokens
    $(S_1, \ S_2)$ and $\delta (\iota(j)-1) + \delta (\iota(j))
    = g_{j+1} - g_j.$
     \end{itemize}

     In the case of commingling in our setting, we merge an $(F_1 , \ F_2)$
     with an $(S_1 , \ S_2)$ to occur in three positions to which we assign
     the tokens $(P_1 , \ P_2, \ P_3)$.  We can use properties of the
     cumulative distribution function of the standard normal probability
     density function to prove

         \noindent {\bf Theorem 7:}
  For a pure Gaussian blur or a mixture of Gaussian blurs with
  $\sigma_{\max} < 0.5 T / \max_j \nu_j$,  if $\iota(j+1) = \iota(j)+2$ and
  $0 < M_{\iota(j)+1,j+1} < 0.5$, then
  $| \delta( \iota(j))| > | \delta ( \iota(j)+1)|$ and
  $| \delta( \iota(j)) + 2| > | \delta ( \iota(j)+1)|$.

  We can now continue our motivating example.

  Recall that
  \begin{eqnarray*}
    \gamma_0 & = & \left( 0 , \ 0 , \ \frac{242}{256} , \ \frac{253}{256} , \
    -1, \ \frac{218}{256} , \ \frac{254}{256} , \ -1, \ -\frac{26}{256} , \
    0, \ 0 \right) \\
 \gamma_1 & = & \left( 0 , \ \frac{5}{256} , \ 1, \ -\frac{209}{256} , \
 -\frac{250}{256} , \ 1, \ -\frac{196}{256} , \ -\frac{254}{256} , \
 0, \   0, \ 0 \right)
  \end{eqnarray*}
  Therefore,
    \begin{eqnarray*}
      \delta_0 & = & \left( 0 , \ 0 , \ \frac{242}{256} , \ \frac{11}{256} , \
      - \frac{509}{256} , \ \frac{474}{256} , \ \frac{36}{256} , \
      -\frac{510}{256} , \ \frac{230}{256} , \ \frac{26}{256} , \
     \ 0 \right) \\
     \gamma_1 & = & \left( 0 , \ \frac{5}{256} , \ \frac{251}{256} , \
     -\frac{465}{256} , \
     -\frac{41}{256} , \ \frac{506}{256} , \
 -\frac{452}{256} , \ -\frac{58}{256} , \ \frac{254}{256} , \
     0, \ 0 \right)
    \end{eqnarray*}

    In cell-library binding in circuit synthesis one also worries about
    labeling elements that affect future labels \cite{d, rudell}; in that
    setting it has proven helpful to consider all possible labels.  In our
    setting we use information from both sequences to work on the joint
    parsing/labeling of each to make inferences about signal amplitude values.
    We begin by seeking the smallest nonzero elements in each difference
    sequence.

    We have $\delta_0 [2] = \frac{242}{256}$ and
    $\delta_1 [1] = \frac{5}{256}$.  Since $\delta_1 [2]$
    has the same sign as $\delta_1 [1]$ and larger amplitude, in
    $\delta_1$ component 1 corresponds to $S_1$ or $A$.
    However, it cannot correspond to $A$ because it is smaller than
    $\delta_0 [2]$.  Therefore, in $\delta_1$ components 1 and 2 are
    parsed as $(S_1, \ S_2)$; therefore,
    $g_1-g_0 = \delta_1 [1] + \delta_1 [2] = 1.$
    Let us return to $\delta_0$.  Since $\delta_0 [2]$ has the same sign
    and larger magnitude than $\delta_0 [3]$, the possible labels for
    component 2 of $\delta_0$ are $A, \ F_1$ or $P_1$.  However, we
    already know that $g_1 - g_0 = 1$, so we can infer that in $\delta_0$
    components 2, 3, and 4 are parsed as $P_1, \ P_2, \ P_3$,  Since
    $\delta_0 [2] + \delta_0 [3] + \delta_0 [4] = -1 = g_2 - g_0$, we
    can infer that $g_2-g_1 = -2$.  We next return to $\delta_1$ to process the first unparsed component, etc.  With this process of going back and forth
    between $\delta_1$ and $\delta_0$, we will parse the components of
    $\delta_0$ as $$(0, \ 0, \ P_1, \ P_2, \ P_3, \ P_1, \ P_2, \ P_3, \
    F_1, \ F_2, \ 0)$$ and the components of $\delta_1$ as
    $$(0 \ S_1, \ S_2, \ P_1, \ P_2, \ P_3, \ P_1, \ P_2, \ P_3, \ 0, \ 0),$$
    infer that $g_3 - g_2 = 2, \ g_4 - g_3 = -2, g_5 - g_4 = 1,$ and use our
    knowledge that $g_0 = 0$ to correctly recover the amplitudes of $g(t)$.
    We next work towards generalizing this approach.

    \section{Counting Arguments for Dataflow}
    In \cite{savari} we used a counting argument to describe the number
    of samples taken between successive discontinuity points, and in
    \cite{s3} we extended this to the number of samples taken between an
    arbitrary pair of discontinuity points.  These papers focus on the case
    with no blur, and in the language of the previous section the parsing of
    the symbols between two successive discontinuity points would be $A$
    followed by some (and possibly no) zeroes.  In the blur regime we
    currently study, there are many more possible behaviors.  The main idea
    is as follows.  A sample taken between $D_j$ and $D_j + \nu_j \sigma_{\max}$ has in the difference sequence the label $F_1$  or $P_1$ depending on other
    entries in the difference sequence.
    A sample taken between $D_j - \nu_j \sigma_{\max}$ and $D_j$ has in the difference sequence the label $S_1$  or $P_2$ depending on other entries in the
    difference sequence.  The remaining labsls $F_2, \ S_2, \ P_3, \ A$ for
    nonzero entries in a difference sequence depend on the context.  Suppose
    $$D_j - D_{j-1} = (n_j - f_j)T,$$
    where $n_j$ is an integer that is at least two and $0 < f_j < 1$.
    For $n_j \geq 3$ there are ten broad categories of parsing possibilities
    for the segment of the difference sequence between $D_{j-1}$ and $D_j$
    which depend on $\nu_{j-1} \sigma_{\max} , \ \nu_{j} \sigma_{\max} , $
    and $f_j T$.  Suppose that the first sample after $D_{j-1}$
    occurs at $D_{j-1} + \Delta_j$ for some $0 < \Delta_j < T$.
    Then within each category there are five possible parsings depending
    on $\Delta_j$.  We can also loosely extend that result to a partial
    description of the beginnings and endings of the parsing for a segment of
    the difference sequence between discontinuity points that are not
    successive.  For $n_j = 2$ and $0 < f_j < 0.5$, the five broad categories
    of parsing possibilities correspond to
    \begin{itemize}
    \item $0 < (1-f_j) T - \sigma_{\max} \nu_j <  \sigma_{\max} \nu_{j-1} <
      (1-f_j) T <  T - \sigma_{\max} \nu_{j-1} $
    \item $0 < \sigma_{\max} \nu_{j-1} < (1-f_j) T - \sigma_{\max} \nu_j <  
      (1-f_j) T <  T - \sigma_{\max} \nu_{j-1} $
    \item $0 < (1-f_j) T - \sigma_{\max} \nu_j <  \sigma_{\max} \nu_{j-1} <
      T - \sigma_{\max} \nu_{j-1} < (1-f_j) T  $      
    \item $0 < \sigma_{\max} \nu_{j-1} < (1-f_j) T - \sigma_{\max} \nu_j <
      T - \sigma_{\max} \nu_{j-1} < (1-f_j) T  $
    \item $0 < \sigma_{\max} \nu_{j-1} <  T - \sigma_{\max} \nu_{j-1} <
      (1-f_j) T - \sigma_{\max} \nu_j <  $            
    \end{itemize}
    Among these categories, only the first and the third will permit possible
    parsings of $P_1, \ P_2$.  To describe how the parsings associated
    with the two cases depends on $\Delta_i$, we will divide each case into
    two.

    Case 1.1: $\sigma_{\max} \nu_{j-1} < f_j T, \
    \sigma_{\max} \nu_{j} < f_j T, \ \sigma_{\max} \nu_{j-1} +
    \sigma_{\max} \nu_{j} > (1- f_j) T$:
    
    \begin{itemize}
    \item $0 <  \Delta_j < (1-f_j) T - \sigma_{\max} \nu_j$: Parse $F_1 F_2$.
      The parsing of the next segment begins $A$.
    \item $(1-f_j) T - \sigma_{\max} \nu_j < \Delta_j <\sigma_{\max} \nu_{j-1}$:
      Parse $P_1 P_2$.
      The parsing of the next segment begins $P_3$.
    \item  $\sigma_{\max} \nu_{j-1} <  \Delta_j < (1-f_j) T$:
      Parse $A S_1 $.
      The parsing of the next segment begins $S_2$.
    \item $(1-f_j) T < \Delta_j < T - \sigma_{\max} \nu_{j-1} $:
      Parse $A$.
      The parsing of the next segment begins $F_1$ or $P_1$.
    \item $T - \sigma_{\max} \nu_{j-1}< \Delta_j < (1-f_j) T
      + \sigma_{\max} \nu_j$:       Parse $S_2$ or $P_3$.
      The parsing of the next segment begins $F_1$ or $P_1$.
    \item  $(1-f_j) T + \sigma_{\max} \nu_{j} < \Delta_j < T$:
      Parse $S_2$ or $P_3$.
      The parsing of the next segment begins $A$.
  \end{itemize}

    Case 1.2: $\sigma_{\max} \nu_{j-1} < f_j T <  \sigma_{\max} \nu_{j}, \
    \sigma_{\max} \nu_{j-1} +
    \sigma_{\max} \nu_{j} > (1- f_j) T$:

        \begin{itemize}
    \item $0 <  \Delta_j < \sigma_{\max} \nu_j - f_j T$: Parse $F_1 F_2$.
      The parsing of the next segment begins $F_1$ or $P_1$.
          \item $\sigma_{\max} \nu_j - f_j T < \Delta_j <
(1-f_j) T - \sigma_{\max} \nu_j$: Parse $F_1 F_2$.
      The parsing of the next segment begins $A$.
\item $(1-f_j) T - \sigma_{\max} \nu_j < \Delta_j < T - \sigma_{\max} \nu_{j-1} $:  Follow Case 1.1.
    \item $T - \sigma_{\max} \nu_{j-1}< \Delta_j < T$:
     Parse $S_2$ or $P_3$.
      The parsing of the next segment begins $F_1$ or $P_1$.
  \end{itemize}
    Case 3.1: $\sigma_{\max} \nu_{j} < f_j T <  \sigma_{\max} \nu_{j-1}, \
    \sigma_{\max} \nu_{j-1} +
    \sigma_{\max} \nu_{j} > (1- f_j) T$:

        \begin{itemize}
        \item $0 < \Delta_j <\sigma_{\max} \nu_{j-1}$: Follow Case 1.1
        \item $\sigma_{\max} \nu_{j-1} < \Delta_j < T - \sigma_{\max} \nu_{j-1}$:       Parse $A S_1 $.
      The parsing of the next segment begins $S_2$.
    \item $T - \sigma_{\max} \nu_{j-1}< \Delta_j < (1-f_j) T$: Parse
      $S_2 S_1$ or $P_3 S_1$. The parsing of next segment begins $S_2$
    \item $(1-f_j) T       < \Delta_j <
      (1-f_j) T + \sigma_{\max} \nu_{j}$
      Parse $S_2$ or $P_3$.
      The parsing of the next segment begins $F_1$ or $P_1$.
    \item $(1-f_j) T + \sigma_{\max} \nu_{j} < \Delta_j < T$:
            Parse $S_2$ or $P_3$.
      The parsing of the next segment begins $A$.
      \end{itemize}

            Case 3.2: $\sigma_{\max} \nu_{j-1} > f_j T, \
    \sigma_{\max} \nu_{j} > f_j T, \ \sigma_{\max} \nu_{j-1} +
    \sigma_{\max} \nu_{j} > (1- f_j) T$:
            \begin{itemize}
            \item $0 <  \Delta_j < (1-f_j) T - \sigma_{\max} \nu_j$: Follow Case 1.2.
            \item $(1-f_j) T - \sigma_{\max} \nu_j <  \Delta_j <
              (1-f_j) T$:                  Follow Case 3.1.
            \item $(1-f_j) T <  \Delta_j < T$:       Parse $S_2$ or $P_3$.
      The parsing of the next segment begins $F_1$ or $P_1$.
      \end{itemize}
   
    \section{On Collaborative Parsing and Amplitude Recovery}
    From \cite[Lemma 2]{savari} we know that the offsets are related by
    $\Delta_j = (\Delta_{j-1} +f_{j-1} T)$ modulo $T$, and
    \cite{savari} and \cite{s3} imply that a refinement of $\Delta_j$
    occurs for each $j$ as more observations about the difference sequence
    are processed.

    In the following discussion we assume that we do not consider identical
    copies of the same difference sequence.

    The sum of the entries of any difference sequence are zero, and
    either $g_{j+1}-g_j$ is the value of a single component of the
    difference sequence, $g_{j+1}-g_j$ is the sum of the values of a
    pair of consecutive components of the difference sequence with the
    same sign, or $g_{j+2}-g_j$ is the sum of the values of three
    consecutive components of the difference sequence.  The nonzero
    components of a difference sequence are partitioned so that each
    component contributes to one of these three cases for exactly one value
    of $j$.

    Our scheme mainly seeks the smallest cluster of components in the
    unprocessed portion of the difference sequence that have a matching sum.
    When we find terms that are identical or have a matching sum, we have
    processed those portions of the difference sequences, we reset the
    algorithm to start searching at the beginning of the unprocessed
    portions of the diffeence sequences.  If we cannot find a match then
    we will be in a situation like our motivating example, and in that
    case there will be collaborative parsing until there is a match in
    the sums.  Those cases are left to the end of the search.
    We initialize by starting at the first nonzero difference value in
    each sequence and we list the searches in order:
    \begin{enumerate}
    \item Do the first elements of the unprocessed portions of the
      difference sequences match?
    \item Does the first element of the unprocessed portion of the
      difference sequence match the sum of the first two components
      of the unprocessed portion of the other difference sequence which
      must be of the same sign?
    \item Is the sum of the first two elements of the unprocessed portions
      of each of the two  difference sequences equal?
    \item Is the unprocessed portion of one difference sequence matching
      a commingling pattern, and are there four consecutive symbols
      in the unprocessed portion of the other which provide a matching sum?
    \item Is the sum of the first three elements of the unprocessed portions
      of each of the difference sequences equal?  This case has multiple parts:
      \begin{enumerate}
      \item If the magnitudes of the component values of one of the sequences
        are in decreasing order, then parse that string as $A, \ F_1, \ F_2$.
\item If the magnitudes of the component values of one of the sequences
are in increasing order, then parse that string as $S_1, \ S_2 , \ A$.
\item If in one of the sequences the second unprocessed component has
  the largest magnitude, then there are again subcases:  If the first and
  third unprocessed components of that sequence have differing signs, then
  the determination of parsing that string as $A, \ F_1, \ F_2$
  or as $S_1, \ S_2 , \ A$ is determined by the sign of the second
  component; i.e., the $A$ is matched to the one component of a
  different sign from the other two.  Otherwise, if the magnitude
  of the first component is larger than the magnitude of the first component
  of the other sequence, then parse as $A, \ F_1, \ F_2$ and if not,
  parse as $S_1, \ S_2 , \ A$.
\item For the remaining subcase, the true parsing of the unprocessed
  portion of one sequence begins as $P_1, \ P_2, \ P_3$ and for the
  other sequence it is either $P_1, \ P_2, \ P_3$ or
  $F_1, \ F_2, \ A$ or $A, \ S_1, \ S_2 $.  As in the previous subcase
  if the first and
  third unprocessed components of that sequence have differing signs, then
  the sign of the second unprocessed component will eliminate a possible
  parsing.  If all three of the first unprocessed components of each sequence
  have the same sign, then as in the previous subcase an argument about
  magnitudes will eliminate a possible parsing.
  For this subcase we know $g_{j+2}-g_j$ for the corresponding value of $j$,
  but we can only bound $g_{j+1}-g_j$ and $g_{j+2}-g_{j+1}.$ Once again,
  the bounds will depend on whether the first and third unprocessed
  components have the same sign or opposite signs.  To resolve this case
  either more information must be gleaned about the values of
  $\Delta_k$ or by using constraints on the relative values of $\nu_j$
  and $\nu_k$, which lead to constraints on the corresponding relative
  values of the magnitudes of the corresponding components of $g_D$.
      \end{enumerate}
    \item The remaining cases extend our motivating example and again
      are divided into subcases.  Note that in the previous section we saw
      that if one sequence is parsed $P_1, \ P_2, \ P_3$ then the other cannot
      be parsed as $ A, \ A$ or as $F_1, \ F_2, \ S_1 , \ S_2$.
      \begin{enumerate}
      \item If the unprocessed portion of the beginning of one sequence
        has the parsing $A, \ P_1 , \ P_2 , \ P_3$, then the third
        component of that segment has smaller magnitude than its second
        or fourth, and its first component has larger magnitude than the
        first unprocessed component of the other sequence.
              \item If the unprocessed portion of the beginning of one sequence
        has the parsing $S_1, \ S_2, \ P_1 , \ P_2 $, then the third
        component of that segment has larger magnitude than its fourth,
        and its first component has smaller magnitude than its second.
                      \item If the unprocessed portion of the beginning of one sequence
        has the parsing $F_1, \ F_2, \ P_1 , \ P_2 $, then the third
        component of that segment has larger magnitude than its fourth,
        and its first two component have the same sign.  If that sign is the
        opposite of the sign of the third component, then the sum of the
        magnitudes of the first two components of that segment is larger
        than the corresponding sum of the other segment.  Otherwise
        the sum of the first three components of this segment is smaller
        than the corresponding sum of the other segment.
        \end{enumerate}
\end{enumerate}

    As a brief comment on stochastic noise, we need enumerative techniques
    or good heuristics to optimize the amplitude recovery problem in the
    presence of underlying logical constraints.

  \section{Bounds on Discontinuity Point Distances}
  Assume that the first sample is taken at $t$.
  We have the following:
  \begin{itemize}
  \item If the segmentation point associated with $D_j$ has the label $P_1$
    or $F_1$,     then
    \begin{displaymath}
t+ \iota (j) T - \nu_j \sigma_{\max}
      < D_j <       t+ \iota (j) T .
    \end{displaymath}
  \item If the segmentation point associated with $D_j$ has the label $S_2$
    or $P_3$,     then
    \begin{displaymath}
      t+ [\iota (j) -1]T
           < D_j <       t+ [\iota (j) -1]T + \nu_j \sigma_{\max} .
          \end{displaymath}
  \end{itemize}
  Therefore, these bounds can be combined to bound $D_{j+k}-D_j$ given
  the labeling of a difference sequence.  Suppose the labels $(x,y)$
  are associated with the segmentation points $(D_j , \ D_{j+k})$.
  \begin{itemize}
  \item If $(x,y) = ( A , \ A)$, then
$
[ \iota (j+k) - \iota (j) -1] T + (\nu_j + \nu_{j+k}) \sigma_{\max}
      < D_{j+k}-D_j <  [ \iota (j+k) - \iota (j) +1] T - (\nu_j + \nu_{j+k}) \sigma_{\max}.$
      \item If $(x,y) = ( A , \ F_1)$ or if $(x,y) = ( A , \ P_1)$, then
        \begin{displaymath}
[ \iota (j+k) - \iota (j) ] T -  \nu_{j+k} \sigma_{\max}
      < D_{j+k} - D_j <  [ \iota (j+k) - \iota (j) +1] T - \nu_j \sigma_{\max}.
        \end{displaymath}
      \item If $(x,y) = ( A , \ S_2)$ or if $(x,y) = ( A , \ P_3)$, then
                \begin{displaymath}
[ \iota (j+k) - \iota (j) -1] T + \nu_j \sigma_{\max}
      < D_{j+k}-D_j <  [ \iota (j+k) - \iota (j) ] T + (\nu_{j-k} - \nu_{j}) \sigma_{\max}.
        \end{displaymath}
              \item If $x \in \{ F_1, \ P_1 \}$  and $y=A$, then
        \begin{displaymath}
[ \iota (j+k) - \iota (j) -1] T +  \nu_{j+k} \sigma_{\max}
      < D_{j+k}-D_j <  [ \iota (j+k) - \iota (j) ] T + (\nu_j - \nu_{j+k}) \sigma_{\max}.
        \end{displaymath}
        \item If $x, y \in \{ F_1, \ P_1 \}$, then
            \begin{displaymath}
[ \iota (j+k) - \iota (j) ] T -  \nu_{j+k} \sigma_{\max}
      < D_{j+k}-D_j <  [ \iota (j+k) - \iota (j) ] T + \nu_j \sigma_{\max}.
            \end{displaymath}
      \item If $x \in \{ F_1, \ P_1 \}$  and $y \in \{S_2 , \ P_3 \}$, then
           \begin{displaymath}
[ \iota (j+k) - \iota (j) -1] T 
      < D_{j+k}-D_j <  [ \iota (j+k) - \iota (j) -1] T + (\nu_j + \nu_{j+k}) \sigma_{\max}.
           \end{displaymath}
         \item If $x \in \{ S_2, \ P_3 \}$  and $y=A$, then
                      \begin{displaymath}
[ \iota (j+k) - \iota (j) ] T + (\nu_{j+k} + \nu_j) \sigma_{\max}.
      < D_{j+k}-D_j <  [ \iota (j+k) - \iota (j) +1] T - \nu_{j+k} \sigma_{\max}.
           \end{displaymath}
                    \item If $x \in \{ S_2, \ P_3 \}$  and $y \in \{ F_1, \ P_1 \}$, then
        \begin{displaymath}
[ \iota (j+k) - \iota (j) +1] T - (\nu_j + \nu_{j+k}) \sigma_{\max}
      < D_{j+k}-D_j <  [ \iota (j+k) - \iota (j) +1] T .
        \end{displaymath}
      \item If $x, y \in \{ S_2, \ P_3 \}$, then
                \begin{displaymath}
[ \iota (j+k) - \iota (j) ] T - \nu_j \sigma_{\max}
      < D_{j+k}-D_j <  [ \iota (j+k) - \iota (j) ] T + \nu_{j+k} \sigma_{\max}.
                \end{displaymath}
  \end{itemize}

  The minimum distance of $1.5T$ often provides tighter lower bounds than
  the ones specified here, but some of the upper bounds constrain the relationship between parsing and amplitude recovery.

\section{Conclusions} 
In {\em \`{A} la recherche du temps perdu}, Marcel Proust suggests that the
  real voyage of discovery consists not in seeking new lands, but seeing
  with new eyes.  Regarding image data, image registration and multiple
  image analysis offer opportunities that single image analysis do not.

\end{document}